# Digital Twin and Artificial Intelligence Incorporated With Surrogate Modeling for Hybrid and Sustainable Energy Systems

Abid Hossain Khan[1], Salauddin Omar[2], Nadia Mushtary[1], Richa Verma[3], Dinesh Kumar[4], Syed Alam[5*]

[1]Institute of Nuclear Power Engineering, Bangladesh University of Engineering and Technology, Dhaka-1000, Bangladesh

[2]Department of Mechanical and Production Engineering, Ahsanullah University of Science and Technology, Dhaka-1208, Bangladesh

[3]Department of Electrical Engineering, Indian Institute of Technology Delhi, Delhi 110016, India

[4]Department of Mechanical Engineering, University of Bristol, Bristol BS8 1TR, UK

[5]Department of Nuclear Engineering and Radiation Science, Missouri University of Science and Technology, Rolla, MO 65409, USA

*Corresponding author: Syed Alam (alams@mst.edu)

**Abstract**

Surrogate modeling has brought about a revolution in computation in the branches of science and engineering. Backed by Artificial Intelligence, a surrogate model can present highly accurate results with a significant reduction in computation time than computer simulation of actual models. Surrogate modeling techniques have found their use in numerous branches of science and engineering, energy system modeling being one of them. Since the idea of hybrid and sustainable energy systems is spreading rapidly in the modern world for the paradigm of the smart energy shift, researchers are exploring the future application of artificial intelligence-based surrogate modeling in analyzing and optimizing hybrid energy systems. One of the promising technologies for assessing applicability for the energy system is the digital twin, which can leverage surrogate modeling. This work presents a comprehensive framework/review on Artificial Intelligence-driven surrogate modeling and its applications with a focus on the digital twin framework and energy systems. The role of machine learning and artificial intelligence in constructing an effective surrogate model is explained. After that, different surrogate models developed for different sustainable energy sources are presented. Finally, digital twin surrogate models and associated uncertainties are described.

**Keywords:** Artificial Intelligence, Hybrid Energy System, Digital Twin, Machine Learning, Surrogate Modeling.

# 1. Introduction

Computer simulation has become an inseparable part of modern science. Although it was first introduced as a tool to investigate the probabilistic nature of nuclear physics and meteorology before World War II, it has quickly found its application in a variety of scientific disciplines (Winsberg, 2013). Whether it is physics, ecology, engineering, or economics, the role of computer simulation is unquestionable. If modeled properly/accurately with perfect physics and models, the obtained results are comparable to real-life experiments in terms of accuracy (Winsberg, 2010). Thus, computer simulations are often utilized as a complementary tool to generate required data for research and development (R&D) instead of conducting expensive experiments (Frenz, 2007). However, a simulation technique itself may be computationally expensive in terms of necessary resources and time-consuming depending on the physical behavior it must imitate. For analyzing the behavior of a system with sufficient reliability, a massive number of simulation runs are required (Frenz, 2007). With the increasing complexity and variability of modern mathematical problems, it is becoming more and more impractical to employ computer simulations directly for analysis of systems with large number of decision variables. Rather, the use of approximating yet accurate data-driven models is identified as a more feasible option for the researchers. Therefore, these "approximate" models, also called "surrogate" models, are becoming more and more popular among the researchers in recent years (Sobester, Forrester and Keane, 2008; Jiang, Zhou and Shao, 2020).

The surrogate model is also known as an "emulator". Surrogate model or emulator is a trained statistical/mathematical model that replaces computer simulations in *predicting the behavior* of a specific system (Sobester, Forrester and Keane, 2008; Jiang, Zhou and Shao, 2020). A surrogate model may be developed to perform analysis related to optimization, risk assessment, multi-criteria decision-making, etc. The main advantage of surrogate models over conventional simulation techniques is that it is much faster than the latter one. Therefore, when the number of variables or inpis are large, they outweigh direct simulations. Surrogate models, on the other hand, are almost entirely dependent on their training process. Machine learning (ML) and artificial intelligence (AI) are used to train a surrogate model. Researchers have found the application of numerous machine learning methods in surrogate modeling (Davis, Cremaschi and Eden, 2018). Depending on whether the surrogate model is local or global, the preference of the training method may vary. There are some methods widely used for local surrogate models such as response surface methodology (RSM) and regression methods (Eason and Cremaschi, 2014; Chelladurai *et al.*, 2021). Kriging and artificial neural networks (ANNs), on the other hand, are commonly used for global surrogate models (Henao and Maravelias, 2010; Schweidtmann and Mitsos, 2019).

Hybrid energy systems (HESs) are believed to be the future of power generation. This emerging concept is expected to bring together different power generation, consumption and storage technologies to form a single, integrated system (Berrada, Loudiyi and El Mrabet, 2021). The benefit of such a system is that it has a wider applicability, higher reliability, and an overall lower cost. Therefore, HESs are also considered as sustainable energy systems (SESs) (Berrada, Loudiyi and El Mrabet, 2021). Although it was originally expected that a HES will be a combination of conventional energy sources with different storage technologies, the recent definitions also consider energy systems as HESs even though the system comprises of renewable sources only (Negi and Mathew, 2014). Some argue that renewable sources are the only energy sources that ensure sustainability of future energy sector (Qazi *et al.*, 2019). Nuclear energy, although not considered as a renewable or sustainable energy source (Pearce, 2012), also exhibits a higher potential to be a part of SES (Ruth *et al.*, 2014; Nowotny *et al.*, 2016). Nuclear-coal hybrid energy systems are being considered by the researchers for hydrogen generation (Chen *et al.*, 2015). In short, the concept of HES is changing with the progression of time. Consequently, there exists wide variabilities in the dynamic behaviors of different SESs and HESs. This makes modeling and analyses of these systems quite challenging (Gupta, Saini and Sharma, 2011; Nag and Sarkar, 2018). Therefore, development of ML-assisted surrogate models for futuristic energy systems has become inevitable (Perera *et al.*, 2019; Ruan *et al.*, 2020).

In this chapter, the concept of surrogate modeling and its characterization process are discussed. Then, the role of AI in the data-driven active learning of a surrogate model is explained. At the same time, the use of different artificial intelligence and machine learning techniques in surrogate modeling is explored. After that, a literature survey is conducted to identify the applicability of surrogate modeling in analyzing and optimizing different hybrid and sustainable energy systems. Finally, a brief discussion on how the concept of surrogate modeling may be implemented in a broader spectrum is provided.

## 2. Surrogate Modeling

Surrogate modeling is the process of constructing a statistical model that are used to bypass expensive numerical simulations and accurately predict the output of the simulations using approximation functions. However, the objective of a surrogate model is not to eliminate computer simulations entirely but to reduce its need as much as possible (Jiang, Zhou and Shao, 2020). To replicate the output of the original simulation, a surrogate model must be trained properly. Training a surrogate model is a machine learning-based data-driven process. In this process, approximate functions are derived from the available data. These approximate functions should be able to generate output data that are in good agreement with the original simulation results. In other words, a surrogate model learns from the data it is provided with and then performs its task such as optimization, risk assessment, sensitivity analysis, etc. based on what it has learnt. Computer simulations are used only to generate the necessary data to train the surrogate model. The approximation functions of a surrogate model are easier to evaluate than the actual differential equations that describe the system. In this way, the computational requirement of a trained statistical model is much lower than that of the original simulation model, and the surrogate models are much cheaper than computer simulations in terms of time-consumption (Davis, Cremaschi and Eden, 2018; Jiang, Zhou and Shao, 2020). Fig.1 presents the basic concept of surrogate modeling and its linkage to computer simulation.

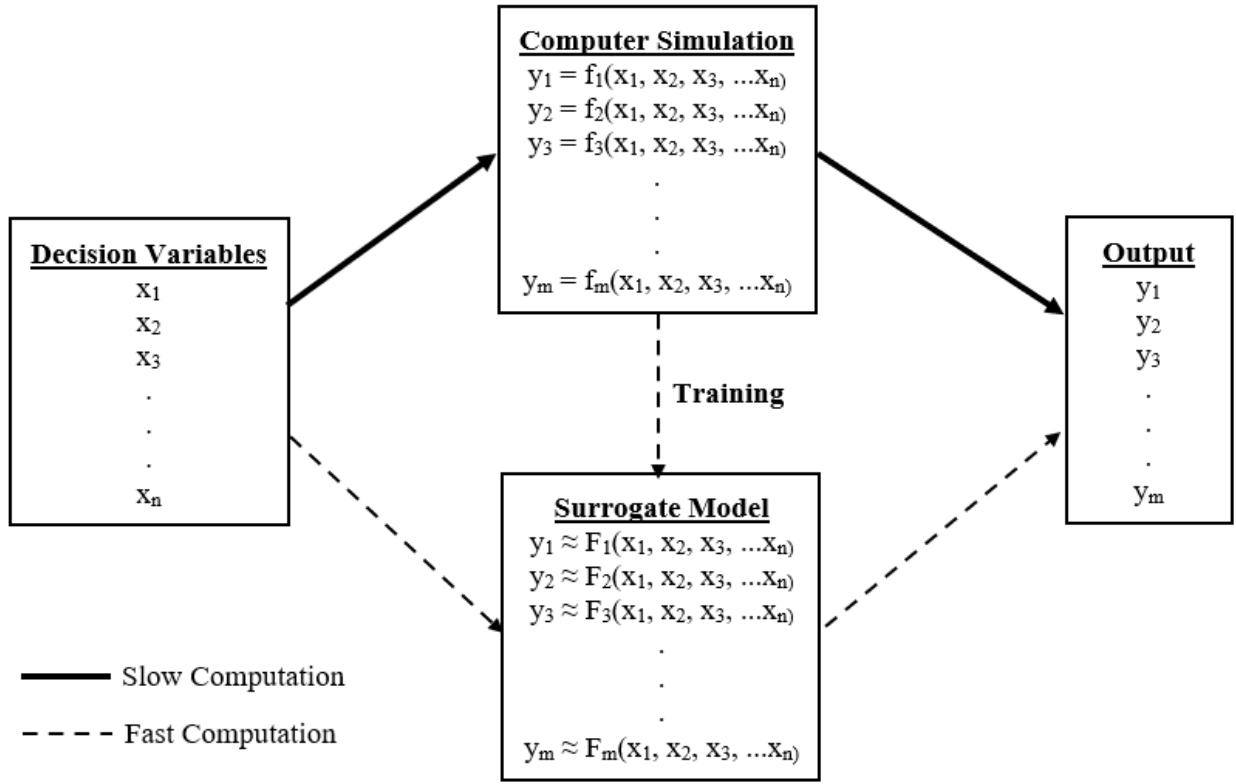

Fig.1. Basic concept of surrogate modeling

From Fig.1, it is clearly understood that a surrogate model replaces the original output functions $f_i(x)$ with approximation functions $F_i(x)$. These approximation functions are data-driven fitting functions. The accuracy of the approximate functions, or how well the model is trained, is directly proportional to the surrogate model's performance. There are two ways to ensure that a model is trained properly. These are to (1) Provide a large sample data for the training; or (2) Employ active learning process to reduce the sampling size. Since it is almost impossible to predict the exact sample size (addressed in Point 1) required to build an accurate surrogate model, active learning (addressed in Point 2) is somewhat preferable (Sobester, Forrester and Keane, 2008). The process flow diagram of surrogate modeling is presented in Fig.2. From Fig.2, it can be observed that an initial sample is selected at the very beginning to generate simulation data for training the surrogate model. After training the model with the simulated data, the approximation functions are used to construct special learning functions that can identify a new sample for further training. With the new sample, new computer simulations are run to generate new training data. Using the enriched dataset, the surrogate model is re-trained. The process is repeated until the surrogate model is sufficiently accurate (Sobester, Forrester and Keane, 2008).

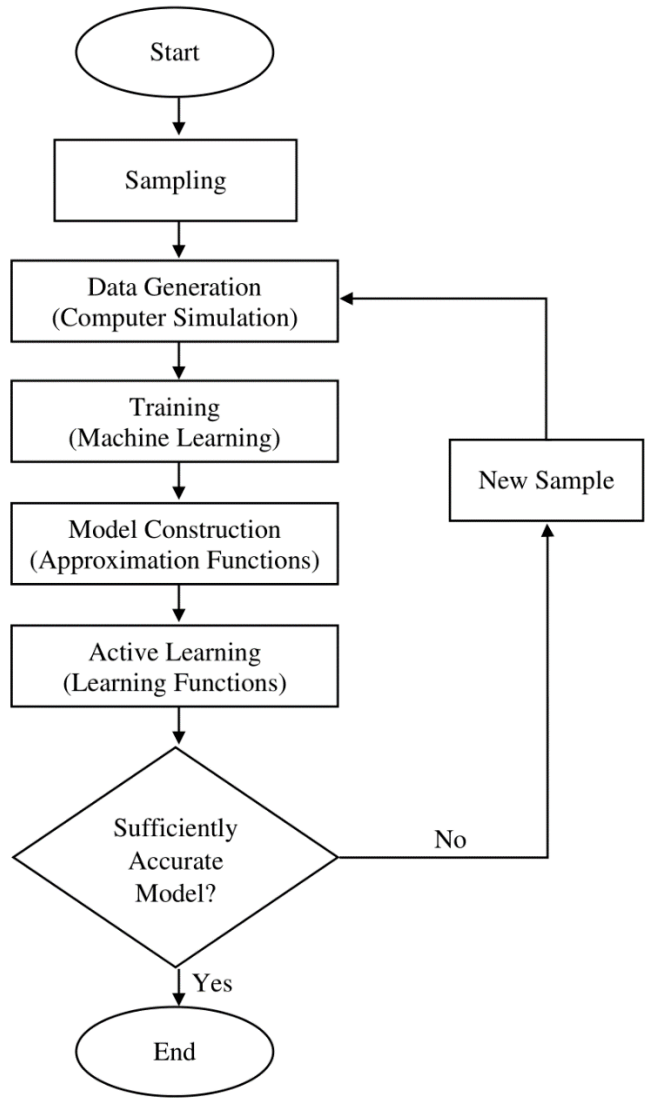

Fig.2. Process flowchart of surrogate modeling

Various computer packages for constructing surrogate models have been developed in a variety of languages. Bouhlel et al.(2019) developed a surrogate modeling toolbox (SMT) (Bartoli, 2019). Other well-known packages are Scikit-learn developed in Python (Pedregosa *et al.*, 2011), SUMO developed in MATLAB (Gorissen *et al.*, 2010), and GPML developed in MATLAB and Octave (Peitz and Dellnitz, 2018). Apart from these surrogate packages, there are numerous surrogate modeling approaches and strategies that are being studied by the researchers. A common surrogate modeling strategy is to consider an ensemble of surrogates to identify the best performing model (Goel *et al.*, 2007). Some researchers have also considered the use of granular method. For example, GA-FGNFN is a surrogate modeling strategy for genetic algorithms. In this strategy, the number of fitness function evaluations is reduced using a granularity technique that takes into consideration the similarity in granule creation between two individuals of the population (Cruz-Vega *et al.*, 2016). Perera et al. (2019) proposed eight different Artificial Neural Network (ANN) architectures to develop a surrogate model which can bypass the Actual

Engineering Model (AEM). The use of self-optimizing principles was proposed by Straus and Skogestad (2018) for efficient development of steady-state surrogate models. Regardless of the modeling strategy, the fundamental goal is to develop a surrogate model as efficiently as possible.

## 3. Artificial Intelligence-Driven Surrogate Modeling

From the previous section, it can be easily realized that surrogate modeling is nothing but a special type of supervised machine learning that is used in engineering and system design. And thus, the training process in surrogate modeling is, to its entirety, within the domain of artificial intelligence. As addressed before, different machine learning approaches are extensively used in surrogate modeling. Some of them are well-established methods such as Kriging (Bouhlel and Martins, 2019), Radial Basis Functions (RBF) (Shan *et al.*, 2009), Inverse Distance Weighting (IDW) (Qiao *et al.*, 2018), Least Squares (LS) (Hastie, Tibshirani and Friedman, 2001), Artificial Neural Network (ANN) (Eason and Cremaschi, 2014), Support Vector Regression (SVR) (Xiang *et al.*, 2017), etc. This section will provide a short overview of these approaches.

### 3.1. Least Squares

The Least Square (LS) Method is a type of mathematical regression analysis (Hastie, Tibshirani and Friedman, 2001). This method is used to find the best fit for the data points by minimizing the sum of squares of the curve's offset points. For example, let us consider a sample dataset ($x_1$, $y_1$),….,($x_{n-1}$, $y_{n-1}$), ($x_n$, $y_n$) where all the x's are the independent variables, and all the y's are the corresponding values of the dependent variables (BYJU'S, 2022). If the approximation function F(x) is the fitting curve obtained from the sample data. Here, the difference between the predicted and actual value at each point is denoted by d (Chakraborty and Bhattacharya, 2013), we may write,

$$d_i = y_i - F(x_i); \quad i = 1, 2, 3 \dots n \tag{1}$$

Thus, the sum of the squares of the differences is,

$$S = \sum_{i \to 1}^{n} d_i^2 = \sum_{i \to 1}^{n}[y_i - F(x_i)]^2 \tag{2}$$

Although this method is fast and simple, it is very accurate for linear problems. However, the method is not suitable for many nonlinear problems (Hastie, Tibshirani and Friedman, 2001).

### 3.2. Inverse Distance Weighting (IDW)

Inverse Distance Weighting (Qiao *et al.*, 2018) is an interpolation method. In order to calculate the values of the dependent variable at the unknown points, known data points are used (Ban and Yamazaki, 2021). Thus, the mathematical expression for the approximation function is,

$$y = F(x) = \begin{cases} \frac{\sum_i^N \beta\left(x, x_k^{(i)}\right) y_k^{(i)}}{\sum_i^N \beta\left(x, x_k^{(i)}\right)}; & x \neq x_k^{(i)} \; for \; all \; i \\ y_k^{(i)}; & x = x_k^{(i)} \; for \; some \; i \end{cases} \tag{3}$$

Here x is the value of the dependent variable at the unknown point, y is the output value at the unknown point, $x_k^{(i)}$ is the value of the dependent variable at the i[th] known point, $y_k^{(i)}$ is the output value for the i[th] known point, N is the number of known points, and β is the weighting function. The simple expression of the weighting function is,

$$\beta = \frac{1}{|x - x_k|^p} \qquad (4)$$

Here p is a positive real number known as power parameter. The value of p is usually taken as two (02), although there is no physical basis to this assumption (Qiao *et al.*, 2018). This method is very simple since no training is required for the model. However, on the downside, it offers an overall poor efficiency, and the maxima and minima of the interpolated curve are always located at the known points (Qiao *et al.*, 2018).

### 3.3. Kriging

Kriging is a Gaussian process regression makes use of prior covariance information to build a more predictive model (Bouhlel and Martins, 2019; Z. Zhang *et al.*, 2021). In this surrogate Kriging modeling approach, the general expression for the approximation function is,

$$F(x) = \sum_{i=1}^{N} \beta_i f_i(x) + \varepsilon(x) \qquad (5)$$

Here the functions $f_i(x)$ are known basis functions, $\beta_i$ are unknown parameters and $\varepsilon(x)$ is a random error function known as correlation component. The term is $\sum_{i=1}^{N} \beta_i f_i(x)$ known as regression component. There are several available options for these components, leading to multiple variants of Kriging (Bouhlel *et al.*, 2016a, Bouhlel *et al.*, 2016b; Bouhlel and Martins, 2019).

Kriging surrogates are very flexible, and construction of these models are less time-consuming. They are also well-suited for high dimensional problems. However, multiple numerical issues arise with this modeling technique when the sample points are too close to each other (Bouhlel and Martins, 2019). Nevertheless, it is one of the most common type of surrogate model (Henao and Maravelias, 2010).

### 3.4. Radial Basis Functions (RBF)

The value of the Radia Basis Function (RBF) function dependent on the distance between a fixed point and the input data point (Shan *et al.*, 2009). If the origin is taken as the fixed point, the function is defined as,

$$Q(x) = Q(\|x\|) \qquad (6)$$

If the fixed point is not the origin, the point is called the center and the function is defined as,

$$Q(x, c) = Q(\|x - c\|) \qquad (7)$$

These functions are subsequently used to construct surrogate models where the approximation function is given by,

$$F(x) = \sum_{i=1}^{N} \lambda_i Q(||x - x_i||) \qquad (8)$$

Here F(x) is the approximation function which sums N radial basis functions. Each function corresponds to different sample point $x_i$ and has a weight of $\lambda_i$ (Shan *et al.*, 2009). Some of the common basis functions are presented in Table 1. This method is simple and fast for small datasets but is vulnerable to oscillations. Points which are too close to each other can pose numerical issues (Shan *et al.*, 2009).

Table 1. Common Basis Functions (r = ||x-x$_i$||)

| Type | Function |
|---|---|
| Linear | $Q(x) = r$ |
| Cubic | $Q(x) = r^3$ |
| Thin Plate Spline | $Q(x) = r^2 \log(r)$ |
| Gaussian | $Q(x) = e^{-\varepsilon r^2}$ |
| Quadratic | $Q(x) = \sqrt{r^2 + \varepsilon^2}$ |

## 3.5. Artificial Neural Network

Artificial Neural Network (ANN), also known as called Neural Network (NN), is a modeling technique inspired by the neural network of a brain (Eason and Cremaschi, 2014). This model imitates the electrical activity of the brain. Among the wide ranges of ML algorithms, ANN is efficient and simple to handle nonlinear systems and robustness. ANN is based on an interconnected collection of artificial neurons named as nodes. Each connection acts like the synapses of a brain and can transmit signals. These signals represent real numbers. The network has three layers, as shown is Fig.3: input layer, hidden (multiple) and output (Mahmood *et al.*, 2020). The output signal from one layer is considered as the input signal of the next adjacent layer. Fig.3 illustrates the basic structure of a 3-layer ANN.

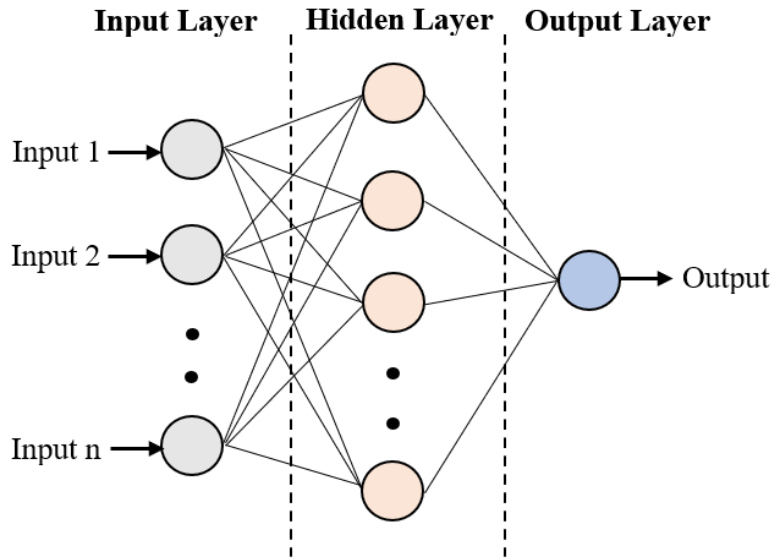

Fig.3. 3-Layer ANN Structure

In ANN, each node has a weight that determines the strength of the transmitted signal. These weights are adjusted as the training of the network is performed. Since the adjustments are not explicit to the end-user, ANN is considered as a black box method (Eason and Cremaschi, 2014). To account for the nonlinearity of the system, activation functions are employed. These functions decide whether to activate or deactivate a particular node of the setwork (Mahmood *et al.*, 2020). Some of the common activation functions are presented in Table 2.  A loss function is employed to determine the performance of the network. Two most common loss functions are Root Mean Square (RMS) error and Absolute Mean Error (AME) (Mahmood *et al.*, 2020).

Table 2. Common Activation Functions in ANN

| Type | Function |
|------|----------|
| Binary | $Bin(x) = \begin{cases} 0 \ for \ x < 0 \\ 1 \ for \ x \geq 0 \end{cases}$ |
| Linear | $Lin(x) = x$ |
| Rectified Linear Unit | $ReLU(x) = \max(0, x)$ |
| Hyperbolic Tangent | $Tanh(x) = \dfrac{2}{1 - e^{-2x}} - 1$ |
| Sigmoid | $Sig(x) = \dfrac{1}{1 - e^{-x}}$ |

ANN is one of the most widely used ML method to construct surrogate models in recent years (Eason and Cremaschi, 2014). ANN has found its application in numerous surrogate models related to risk assessment (Yoon *et al.*, 2020), waste management (Cho *et al.*, 2021), disaster modeling (Kim *et al.*, 2015), design optimization (Sunny *et al.*, 2013; X. Zhang *et al.*, 2021), performance assessment (Le and Caracoglia, 2020), etc. To minimize the number of trials and, subsequently, the time necessary for the surrogate modeling process, ANN along with MEMO (Multimodal-based Evolutionary Multiple Objective) have been proposed (Tutum and Deb, 2015). Novel sampling techniques are also being investigated (Eason and Cremaschi, 2014). The major limitation of this method is the requirement of a large dataset for training the network (Eason and Cremaschi, 2014).

### 3.6. Support Vector Regression

Like ANN, Support Vector Regression (SVR) is also a black box method (Shi *et al.*, 2020, Brereton and Lloyd, 2010). The mathematical expression is,

$$F(x) = \sum_{i=1}^{N} w_i \psi(x, x_i) + \mu \tag{9}$$

The SVR surrogate is very similar to RBF and Kriging surrogates except for the method of estimating the unknown parameters. The objective of SVR is to minimize the value of the following expression,

$$M(x) = \frac{1}{2}|\boldsymbol{w}|^2 + C \sum_{i=1}^{n} |\xi_i| \tag{10}$$

Subjected to the constraint given by,

$$|y_i - w_i \psi(x, x_i)| \leq \varepsilon + |\xi_i| \tag{11}$$

Here $\varepsilon$ is the acceptable error, $\xi_i$ is the slack variable and C is a pre-defined constant controlling the accuracy of the model. SVR can be used to develop surrogate models for both linear and nonlinear systems (Viana *et al.*, 2012; Bourinet, 2016; Shi *et al.*, 2020). For high-dimensional models with a high nonlinearity, SVR works better than many ML techniques (Alizadeh, Allen and Mistree, 2020). However, this method is very time-consuming because of the complexity in calculating the unknown parameters (Bourinet, 2016; Tsirikoglou *et al.*, 2017; Xiang *et al.*, 2017). So, the use of SVR is a tradeoff between high dimensionality and accuracy. Initial versions of SVR were slow whereas the recently developed versions are much faster than before (Xiang *et al.*, 2017).

## 4. Application in Hybrid and Sustainable Energy Systems

Energy demands have risen dramatically with the fast expansion of the global economy, particularly in emerging economies. The fact that fossil fuel supplies needed for energy generation are finite in nature, as well as it is linked to the environmental pollution, a renewed interest in conserving energy and protecting the environment is being observed in recent years (Vine, 2008). One of the strategies to tackle this remarkable growth in energy requirement and ensuring environmental preservation is to harness the power of sustainable and hybrid energy. The development of such energy systems can be achieved successfully by addressing some of the key challenges such as energy saving, system efficiency enhancement for higher productivity and replacement of the conventional fossil fuel power sources with clean energy (Lund, 2007). These challenges can only be dealt with if accurate modeling techniques to predict the dynamics of global energy systems are in hand. This is the reason why modeling of energy systems has gained so much attention of the researchers (Lopion *et al.*, 2018). Researchers have introduced and analyzed different types of energy models to solve various complexities in the field of power production. These models may be used to perform a wide variety of tasks such as forecasting, optimization, waste management, etc. A categorical classification of energy models has been shown in Fig.4.

Hybrid and sustainable energy systems are expected to take over the whole energy network in near future (Berrada, Loudiyi and El Mrabet, 2021). Although hybrid energy systems are still at the initial stage of development, the potential of HESs is immense. The focus is slowly shifting towards HESs that consist of only renewable energy sources (Negi and Mathew, 2014), but there are researchers who believe that other conventional energy sources can't be completely replaced, at least not in near future. To ensure sustainability, inclusion of nuclear and coal-fired energy in HES planning is being recommended (Ruth *et al.*, 2014; Chen *et al.*, 2015). With so many possible directions to go on, effective modeling techniques for HESs are constantly being explored by the scientific community.

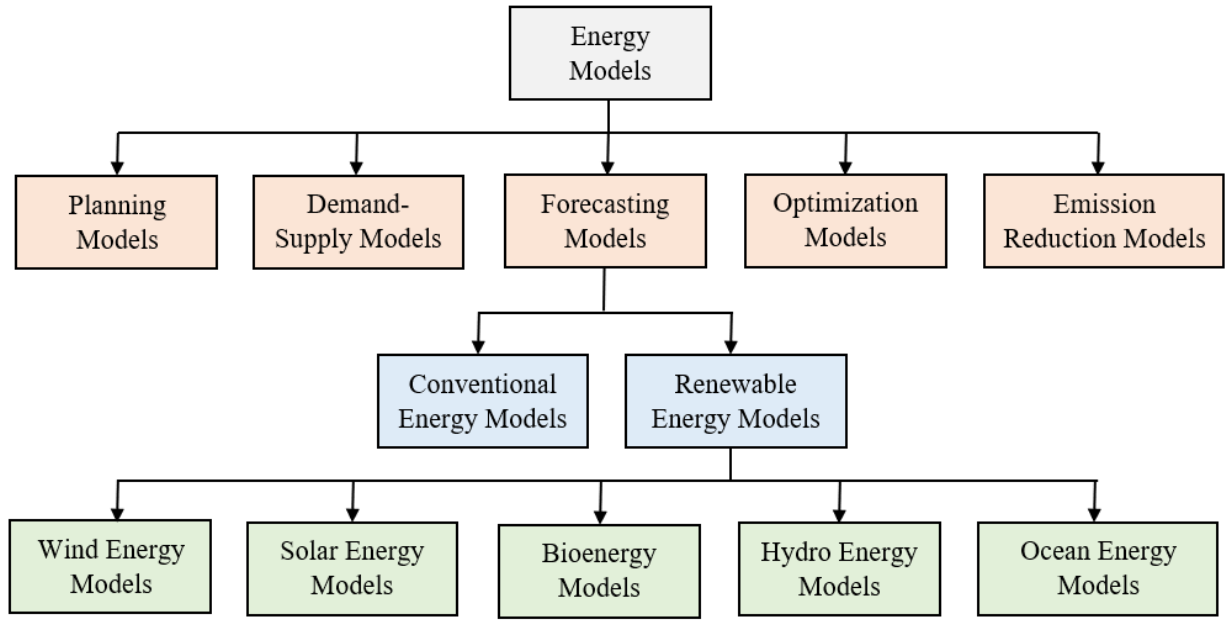

Fig.4. Classification of energy models

Computational optimization has become extremely popular among the researchers for designing different components of an energy system with improved performance and economic feasibility. From 1989 to 2009, there was an exponential increase in research progress that centered around computational optimization methodologies and the development and implementation of novel algorithms for design and analyses (Baños *et al.*, 2011). In recent times, surrogate models are gradually becoming a preferable tool for framework development as it they can reduce the computational expenses significantly while ensuring sufficient accuracy (Evins, 2013). A surrogate model that is trained by ML and AI may be utilized to optimize an energy system with reduced computational time compared to an Actual Engineering Model (AEM), which employed surrogate models for energy system optimization research so vast and recurring (Jiang, Zhou and Shao, 2020). Table 3 presents a review of some recent works that have been conducted to improve different sustainable energy technologies with the help of AI-driven surrogate models.

Table 3. Summary of recent research on sustainable energy system optimization using AI-driven Surrogate Models

| Author(s) | Research Outline | Key Findings |
|---|---|---|
| (Ezhilsabareesh, Rhee and Samad, 2017) | Optimization of a bidirectional impulse turbine's shape | 1. Multiple Surrogate-based multi-objective algorithms combined with CFD simulations may be used to increase bidirectional impulse turbine efficiency |
| (Perera *et al.*, 2017) | Optimization of energy hub distribution | 1. A new hybrid surrogate model optimization framework<br><br>2. Significant reduction in computational |

| | | expenses is achieved |
|---|---|---|
| (Starke *et al.*, 2018) | Optimization of hybrid CSP+PV system | 1. Optimized PTC+PV plants provide a desirable LCOEs with highest capacity factors |
| (Mahulja, Larsen and Elham, 2018) | Surrogate-based framework development for design optimization of wind farms | 1. an optimum design was achieved by almost tripling the investment while compromising wind farm performance<br><br>2. Wake loss effect and fatigue degradation are interrelated<br><br>3. Optimality heavily depends on market value predictions |
| (Sun *et al.*, 2018) | Probabilistic method for wind forecasting | 1. The proposed forecasting method, in comparison to the baseline methods, can mitigates 35% of the pinball loss |
| (Kaya and Hajimirza, 2018a) | Design optimization of organic solar cells | 1. Optimization may lead to 325% increase in absorption. |
| (Ju and Liu, 2019) | Optimization of Wind farm layout | 1. Under similar wind distribution scenario, larger wind farms have shown higher conversion efficiency. |
| (Sun *et al.*, 2019) | Optimization for probabilistic wind forecasting | 1)Higher pinball loss metric score than a baseline quantile regression forecasting model is achieved. |
| (Radaideh and Kozlowski, 2019) | Framework development for advanced nuclear energy modeling | 1. Delayed neutron fraction exhibited larger uncertainties compared to the decay constants. |
| (Zhou and Zheng, 2020) | Uncertainty optimization for materials used in renewable systems | 1. The use of an uncertainty-based optimization technique can help to boost peak power and electricity output. |
| (Barlas *et al.*, 2021) | Design optimization of wind turbine | 1. Surrogate-based optimization framework allows a design that attains up to 6% of load-neutral gain in yearly energy generation. |

From Table 4, it may be observed that AI-driven surrogate models have been utilized in optimizing numerous sustainable energy system designs. One the most investigated sustainable energy source is wind energy because of being environment friendly and cost-efficient. To achieve high demand using wind power, wind farms are being studied in terms of their efficiencies, optimum location and layout, and economic expenses (Mahulja, Larsen and Elham, 2018; Ju and Liu, 2019). At the same time, surrogate model-based optimizations of wind turbines are also being performed (Barlas *et al.*, 2021). Finally, to plan and operate power systems based on wind energy, a better wind forecasting may contribute to a more improved power output. Surrogate models are being employed to predict wind speed distributions in a specific area (Sun *et al.*, 2018, 2019).

Rapid optimization is critical for solar power technology and surrogate models may play a significant role in optimizing the efficiency. A surrogate based Neural Network architecture may accelerate the optimization of thin film solar cells designs (Kaya and Hajimirza, 2018b). An increasingly popular photovoltaic device, organic solar cell (OSC) is now being considered as a potential candidate for future energy systems because of the convenience in fabrication process, low power-cost, and ease in handling. The design optimization based on neural network trained surrogate analysis can result in performance enhancement of OSCs (Kaya and Hajimirza, 2018a).

The combination of concentrated solar power (CSP) and thermochemical energy storage (TCES) also offers a form of renewable energy that is both cost-efficient and ready-to-deploy. Fixed-bed reactors that operate in a CSP plant can be re-imagined as a surrogate model and simulated using realistic dimensional data and algorithms for process optimization (Peng *et al.*, 2020). To reduce the expenses for operations and installation as well as a consequential increment of the capacity factor, hybridization of Concentrating Photovoltaic (CSP) and Photovoltaic (PV) systems have proven to be very promising. The process of hybridization can be executed by the application of a surrogate model that will be trained with genetic algorithm (GA). With such an optimization routine, performance maps may be generated that act as an indication of the economic and operational feasibility of the system (Starke *et al.*, 2018). Ocean waves are an unfathomable source of renewable energy. However, the unpredictability of ocean waves makes it a heavy task to harness such a natural energy to produce electricity. Surrogate driven multi-objective algorithms are being employed in optimization of bidirectional turbines of impulse type to ensure maximum harvesting of this form of energy (Ezhilsabareesh, Rhee and Samad, 2017).

Another important source of energy which is believed to be a part of future sustainable energy systems is nuclear energy (Ruth *et al.*, 2014). Since nuclear power is a baseload energy source with low carbon footprint, it will not be out of the picture soon. Surrogate modeling can be utilized to assist a global sensitivity analysis (GSA) for a nuclear reactor assembly (Banyay, Shields and Brigham, 2019). Uncertainty quantification is another arena where artificial intelligence and surrogate models are proven to be powerful tools for gaining a clearer understanding of the physical phenomenon that occur inside a nuclear reactor (Radaideh and Kozlowski, 2019). This facilitates the amelioration of reactor performance. Nuclear data extracted from real-life reactors combined with multi-physics coupling and simulations and artificially trained surrogate models can operate in conducting uncertainty analysis and design optimization for nuclear reactors (Radaideh and Kozlowski, 2019).

Apart from design optimization of energy systems, there may be other applications of surrogate modeling in hybrid and sustainable energy systems. Surrogate models may be utilized to develop multi-layer uncertainty-based optimization frameworks for renewable energy systems (Zhou and Zheng, 2020). The concept of energy hub can also be benefited by surrogate modeling. However, the integration of optimal dispatch and energy system size is a challenge which makes designing of well-planned and optimally distributed energy hubs a difficult endeavor. Surrogate models coupled with optimization algorithms are proven to be very effective in overcoming these problems (Perera *et al.*, 2017). Surrogate-based P2P market framework development may also be linked to hybrid energy community data to establish a market framework that ensures data privacy. Such frameworks may contribute to the reduction energy expenses while providing financial advantages to community agents (Ju and Liu, 2019). Uncertainty quantification and surrogate model have been developed by authors as well for nuclear energy application (Kumar, Alam, Vučinić, *et al.*, 2020; Kumar *et al.*, 2021, 2022). Different nuclear data adjustment methods have been suggested (Kumar *et al.*, 2019; Kumar, Alam, Sjöstrand, *et al.*, 2020). In short, the scope of surrogate modeling in hybrid and sustainable energy systems is quite vast and versatile.

After a careful observation of Table 4, it may be realized that most of the recent research that utilized surrogate models are focused on optimization of a single sustainable energy source. Apart from a few exceptions where two completely different technologies have been coupled to the energy system (Starke *et al.*, 2018; Peng *et al.*, 2020), none of the studies in the available literature has investigated an energy system with more than two different renewable energy technologies using a surrogate model. To the authors' knowledge, most of the studies on the modeling for HESs have employed direct computer simulation-based methods (Gupta, Saini and Sharma, 2011; Nag and Sarkar, 2018). The reason behind this research gap is the unavailability of the valid experimental datasets. Since the concept of HES is somewhat new, there are very few real-life data to support the simulation results of the proposed models for these systems. As a result, the uncertainties associated with the accuracy of these simulation results is quite high (Torregrossa *et al.*, 2016; Larsen *et al.*, 2019; Cevasco, Koukoura and Kolios, 2021). To exploit the computational advantages of surrogate modeling, result validation as well as sufficient data generation with available simulation models are required.

## 5. Digital Twin Surrogate Models & Associated Uncertainties

In order to ensure practical implementation of the digital twin for HES, the development of this surrogate model would be a major step forward. For the Digital Twin system, the authors are working on a new tool for surrogate modeling according to the proposed system by General Atomics Electromagnetic Systems for the DOE Nuclear Technologies (Jacobsen, 2022). The tool serves as a foundation for advancing the implementation of the HES's various components. A key component of the proposed work is a collaborative effort with the General Electric (GE) team to develop digital twin framework using ML/AI-driven surrogate model for energy system framework. It needs to be accurate and validation should be performed leveraging commercially available models. The goal of this project is to create a "virtual twin" of HES. According to the DOE proposal of General Atomics Electromagnetic Systems in collaboration with INL and LANL (Jacobsen, 2022), we are developing digital twin framework following the route:

- ML/AI-based surrogate model to develop the digital twin system, while using simplifying equations to keep the model as simple as possible to understand.
- By incorporating relevant data and incorporating it into existing energy models, we can perform surrogate model validation.
- Integrate HES surrogate into the existing energy model by expanding it to include key HES energy system behavior.

- Use system and experimental data and perform robustness analysis under uncertainties.
- Begin by incorporating and demonstrating its first use in a real-world energy system.

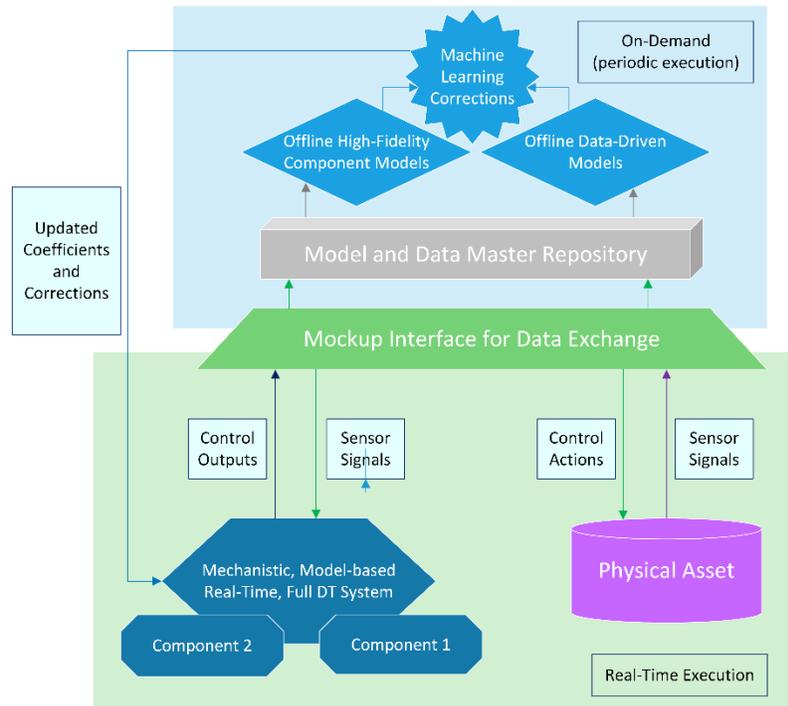

Fig.5. Framework for interactions between digital twin and physical assets (Kochunas and Huan, 2021)

For digital twin, there is an abundance of technology that can be used today. Here is an illustration of how Digital twin work in real time, as shown in Fig.5. They are based on mechanistic models and use data from a physical asset to predict how the digital twin will respond and then take appropriate control actions. This is happening right now. An FMI transitions into an area where on-demand execution is possible. A repository of data is used to keep track of all of the operations. Machine learning processes will be used in this area in an effort to analyze the dynamics model and make adjustments to the digital twin, which will be repeated (Kochunas and Huan, 2021).

Developed digital Twin concept needs to be robust against uncertainty in data and therefore, we propose machine learning-based uncertainty quantification (UQ) and sensitivity analysis (SA). Establishing digital twin's reliability and trustworthiness is critical for their implementation in practice, particularly in safety/mission-critical settings with the potential for catastrophic consequences (Kochunas and Huan, 2021). UQ is a key enabler for assessing these traits. Designers have access to information about the digital twin's levels of confidence and uncertainty can learn about the various responses and outcomes that can be expected. Consequently, it is possible to make well-informed decisions about the control, design, policy or further experiment in question. Since the digital twin's uncertainty is critical for decision support systems, UQ plays an important role. There are many different types of computer science and engineering that use the VVUQ system (Oberkampf, Trucano and Hirsch, 2004; Kochunas and Huan, 2021), which is a way to verify and validate code and models. This system has become the standard in many of these fields. Incorporating uncertainty quantification is a difficult task for digital twin's (UQ). The authors already developed machine learning based UQ (sparse polynomial chaos expansion) and SA (Sobol' indices-based global SA) systems (Kumar, Alam, Vučinić, *et al.*, 2020; Kumar, Koutsawa, Rauchs, *et al.*, 2020; Kumar *et al.*, 2021, 2022), as shown in Fig.6, which can be incorporated in digital twin system. Authors also propose to implement the Forward UQ, Inverse UQ and Optimization under uncertainty, as recommended by the dedicated study on digital twin concepts with uncertainty (Kochunas and Huan, 2021) in their upcoming studies.

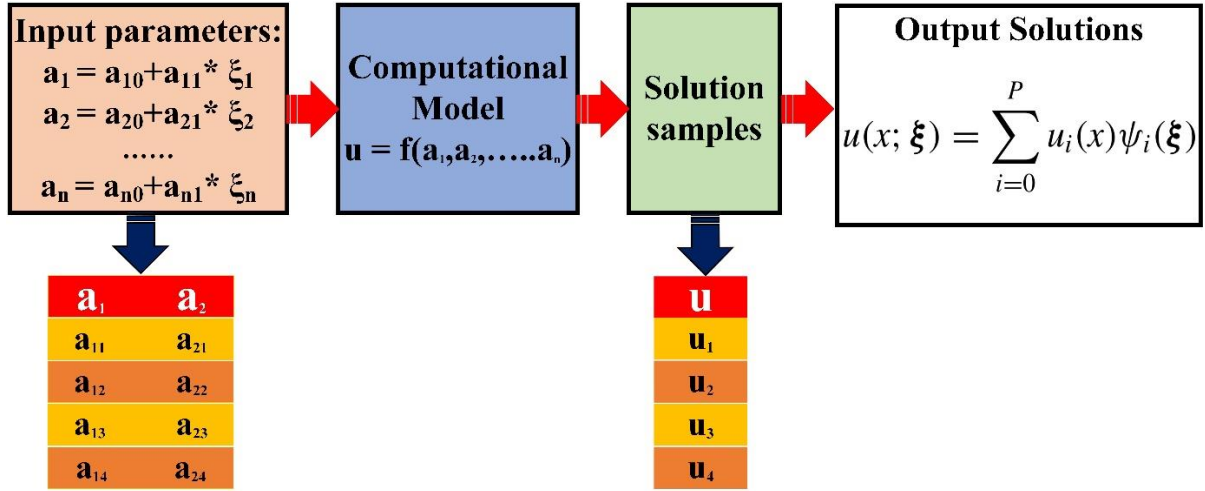

Fig.6. Developed UQ and SA Method

Also, as suggested by Chakraborty, Adhikari and Ganguli (2021), we propose to employ a couple of techniques and extend the directions of digital twin research:

- Developing digital twin model for systems purely on available data, where physics will be discovered and understood from the data using AI algorithms.
- Deep learning algorithms can be employed within digital twin framework in the multiphysics environment.
- A digital twin's primary function is to make predictions about the performance; however, physics-based digital twin performs poorly in this area (Ganguli and Adhikari, 2020). For short-term responses, however, the Gaussian Process-based digital twin model is able to predict. Developing ML/AI-based surrogate-driven digital twin framwrok that can predict how the system will respond over the long term.
- As recommended by Chakraborty, Adhikari and Ganguli (2021), we propose to extend digital twin applications for evaluating the accident conditions using hybrid models for digital twins. hybrid models are expected to outperform single digital twin surrogates.

**Conclusion**

Surrogate models are accurate yet computationally less expensive substitute of computer simulations. For a system with many decision variables, the computational expenses of computer simulations become so high that repetitive simulation runs become somewhat impractical.

Surrogate models can handle this kind of situation with ease. These models construct approximation functions with sample dataset with the assistance of ML and AI-based training process. The approximation functions can give output values close to the simulation results.

This work attempts to identify different AI-assisted surrogate modeling techniques and their functionality for energy systems with specific application of digital twin and associated uncertainties. From the literature survey presented in this work, it may be realized that numerous machine learning method may be utilized to train a surrogate model. The performance of a surrogate model is entirely dependent on how it has been trained and how well it is trained. Each ML method has its advantages and drawbacks; the choice depends on the physical nature of the system to be modelled. Nevertheless, Kriging surrogates and ANN surrogates are the two most common global models.

This work also explored the applicability of surrogate modeling for hybrid and sustainable energy systems. A careful analysis of the available literature suggested that the researchers in the field of energy engineering already used this powerful tool to optimize different conventional and renewable energy sources. At the same time, surrogate models are being developed for small-scale hybrid energy systems consisting of renewable energy sources incorporating digital twin framework. However, the number of studies dealing with a large-scale hybrid energy system is quite rare. This may have resulted from the lack of sufficient simulation data for hybrid energy systems since the concept has emerged just recently. Therefore, emphasis should be given on data generation for hybrid energy systems so that they may be handled effectively through surrogate modeling.